\documentclass[]{ceurart}

\usepackage{booktabs}
\usepackage{listings}
\usepackage{graphicx}
\usepackage{tabularx}
\graphicspath{{figures/}}

\begin{document}

\copyrightyear{2026}
\copyrightclause{Copyright for this paper by its authors.
  Use permitted under Creative Commons License Attribution 4.0
  International (CC BY 4.0).}

\conference{CLEF 2026 Working Notes, 21--24 September 2026, Jena, Germany}

\title{DS@GT-ARC at eRisk 2026 Task 3: Sparse, Semantic, and LLM Reranking for ADHD Symptom Sentences}
\title[mode=sub]{Notebook for the eRisk Lab at CLEF 2026}

\author[1]{David Guecha}[orcid=0009-0009-9855-5330,
email=dahumada3@gatech.edu
]

\address[1]{Georgia Institute of Technology, North Ave NW, Atlanta, GA 30332}
\cortext[1]{Corresponding author.}

\begin{abstract}
This paper describes our submissions to eRisk 2026 Task 3, ADHD Symptom Sentence Ranking. The task requires systems to rank candidate Reddit sentences according to their relevance to each of the 18 symptoms in the Adult ADHD Self-Report Scale (ASRS-v1.1). Because no annotated training data were released for this first edition of the task, we relied on zero-shot experimentation, manual validation, and unsupervised or weakly guided retrieval pipelines. Our systems combine sparse BM25 retrieval, evidence-aware rescoring for self-referential symptom reports, embedding-based reranking, query-prototype expansion, and LLM-based reranking. All submitted systems follow a staged retrieval design in which BM25 retrieves candidates at scale and semantic or LLM rerankers refine the final rankings. Among our submissions, the LLM reranker achieved the strongest official scores, followed by the prototype query-expansion run. Our manual top-10 analysis aligned with the official expert scoring trend, suggesting that staged reranking is a promising direction for further development.
\end{abstract}

\begin{keywords}
  eRisk \sep
  ADHD \sep
  symptom ranking \sep
  information retrieval \sep
  reranking \sep
  large language models
\end{keywords}

\maketitle

\section{Introduction}

Attention-deficit/hyperactivity disorder (ADHD) is a common neurodevelopmental condition whose effects often persist into adolescence and adulthood. It is also strongly intertwined with everyday online activity: recent work has reported associations between ADHD symptoms and problematic social-media use, while emphasizing the need for careful interpretation of these signals~\citep{dekker2023adhdSocialMedia}. Social media therefore contains large quantities of self-authored language about attention, impulsivity, disorganization, restlessness, and related functional difficulties. Retrieval systems that can surface relevant symptom evidence may help researchers and mental-health professionals build better resources for prevention and support, provided that such systems are used for assistive analysis rather than covert profiling.

eRisk 2026 Task 3 introduces an ADHD symptom sentence ranking task built around the 18 symptoms in the Adult ADHD Self-Report Scale (ASRS-v1.1)~\citep{kessler2005asrs,PerezEtAl2026eRiskWorkingNotes,PerezEtAl2026eRiskLNCS}. For each ASRS symptom, participants submit a ranked list of up to 1000 candidate sentences ordered by estimated relevance. A sentence is considered relevant when it conveys information about the author's state with respect to the target symptom, regardless of whether the sentence affirms or denies the symptom. This definition makes the task more clinically meaningful than keyword matching: a good system must identify evidence about the user's behavior, not merely sentences that contain symptom-related terms.

The released Task 3 collection is sentence-tagged and TREC formatted, with each candidate represented by a sentence identifier, optional left context, sentence text, and optional right context. Our local converted parquet collection contains 4,170,875 sentence rows with \texttt{DOCNO}, \texttt{PRE}, \texttt{TEXT}, and \texttt{POST} fields. We treat each ASRS symptom as a separate information retrieval query and generate one ranking per symptom.

\begin{table}[t]
  \caption{Collection statistics from the locally converted Task 3 sentence collection.}
  \label{tab:collection-stats}
  \centering
  \begin{tabular}{lr}
    \toprule
    Collection statistic & Value \\
    \midrule
    Per-user parquet files & 4,521 \\
    Sentence rows & 4,170,875 \\
    Non-empty sentence rows & 4,170,875 \\
    Mean sentences per user file & 922.6 \\
    Median sentences per user file & 624 \\
    Mean sentence length & 13.0 words \\
    Median sentence length & 11 words \\
    \bottomrule
  \end{tabular}
\end{table}

To study this setting, we combine classical IR methods with contemporary semantic and API-based LLM reranking approaches. Our submissions range from lexical BM25 retrieval to semantic embedding reranking and LLM reranking. The central question is whether classical IR methods can produce useful symptom-relevant rankings in a zero-shot setting, and whether semantic or LLM rerankers improve relevance over the sparse retrieval baseline. This task is closely related to previous eRisk symptom-ranking work on depression, particularly sentence ranking for Beck Depression Inventory-II (BDI-II) symptoms~\citep{erisk2023,perez2023bdisen,perez2025depresym}. The ADHD setting differs in two important ways. First, this is the first edition of the ASRS ranking task, so no annotated ADHD training data were provided. Second, ADHD-related symptom evidence in social media may be expressed indirectly, casually, or through everyday functional difficulties, making sparse lexical matching especially brittle. For example, ambiguous terms such as ``turn'' can retrieve non-symptom uses even when they overlap with an ASRS item: ``I got impatient so I took a left turn'' is not evidence for difficulty waiting one's turn, while ``I was impatient waiting for my turn so I left'' is much closer to the target symptom.

\section{Related Work}

The eRisk lab has previously organized sentence-level symptom ranking tasks for depression based on the BDI-II questionnaire. In eRisk 2023 Task 1, participants ranked sentences for 21 depression symptoms, and relevance judgments were later created through pooling and expert assessment~\citep{erisk2023}. The resulting BDI-oriented resources, including BDI-Sen and DepreSym, frame symptom relevance as evidence about the writer's state rather than mere topical relatedness~\citep{perez2023bdisen,perez2025depresym}.

Recent eRisk 2025 submissions to the BDI-II sentence-ranking task explored methods that translate directly to the ASRS setting, including fine-tuned and prompt-based approaches, lexical and semantic retrieval, first-person filtering, reranking, and LLM-assisted symptom matching~\citep{inescid2025erisk,hulat2025erisk,lhs7122025erisk,cotecmar2025erisk,sonuit2025erisk,semanticRetrieval2025erisk,llmPersonas2025erisk}. Unlike many prior depression-ranking submissions, our setting has no ADHD-specific training labels, so we focus on zero-shot ranking and compare how much relevance can be gained through reranking over a shared sparse-retrieval candidate pool. This prior depression work motivates two choices in our ADHD systems. First, we emphasize first-person and self-report language because assessor guidelines for prior BDI tasks distinguish sentences about the writer from generic or third-party discussion. Second, we use multistage retrieval: a sparse first stage supplies candidate recall, and later rerankers attempt to recover semantic and clinical relevance that is often missed by exact lexical overlap.

\section{Task Setup}

For each of the 18 ASRS symptoms, systems submit a TREC-style ranking:

\begin{lstlisting}
symptom_number Q0 sentence-id position_in_ranking score system_name
\end{lstlisting}

We submitted five runs. All submitted files contain 18 rankings with 1000 sentences per symptom. This format emphasizes ranking quality across a large candidate pool rather than binary symptom classification. The LLM source run ranked 2000 candidates per symptom, and the submitted copy was truncated to the top 1000 per symptom.

The ASRS questionnaire contains 18 Likert-style self-report items. The first six items form the widely used screener subset, while items 7--18 provide additional symptom-specific context. We frame each questionnaire item as an information retrieval query: the system must retrieve sentences that provide evidence about the author's state with respect to that item, not answers to the questionnaire itself.

\section{Methodology}

\begin{figure*}[t]
  \centering
  \includegraphics[width=.96\linewidth]{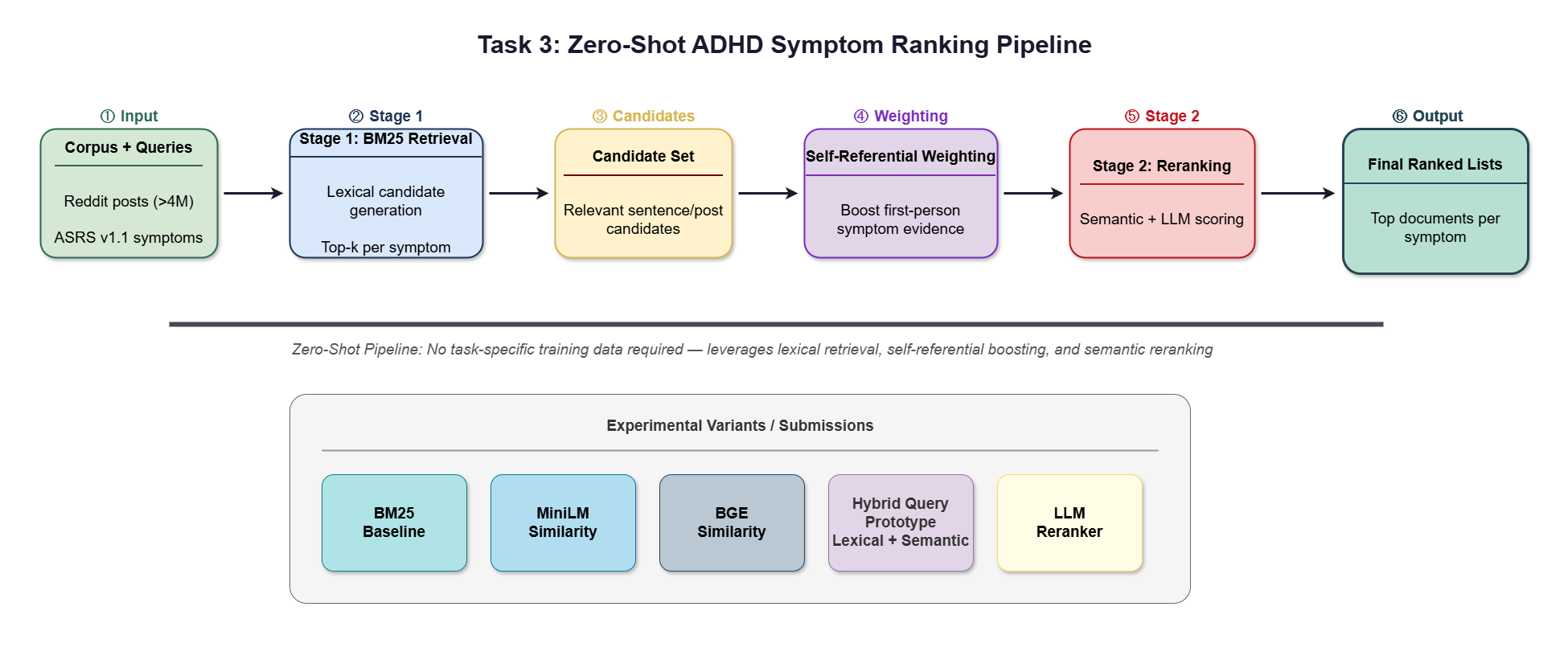}
  \caption{System overview for the submitted Task 3 retrieval and reranking pipelines.}
  \label{fig:system-overview}
\end{figure*}

Figure~\ref{fig:system-overview} summarizes the five submitted pipelines. Each system starts from the ASRS symptom descriptions and produces one ranked sentence list per symptom, with later runs adding semantic or LLM-based reranking on top of sparse retrieval candidates. The submitted systems are summarized in Table~\ref{tab:runs}.

\begin{table}[tbp]
  \caption{Submitted runs for eRisk 2026 Task 3.}
  \label{tab:runs}
  \centering
  \small
  \begin{tabularx}{\linewidth}{l l X}
    \toprule
    Submission & Method & Summary \\
    \midrule
    \texttt{run1} & BM25 v2 & BM25 with noise filtering and evidence-aware rescoring \\
    \texttt{run2} & MiniLM & MiniLM cosine reranking over BM25 candidates \\
    \texttt{run3} & BGE-small & BGE-small cosine reranking over BM25 v2 candidates \\
    \texttt{run4} & Prototype MiniLM & MiniLM reranking with hybrid ASRS/Likert query prototypes \\
    \texttt{run5} & Gemini LLM & Gemini LLM reranking over BM25 v2 candidates \\
    \bottomrule
  \end{tabularx}
\end{table}

The five runs were designed to compare increasingly expensive retrieval and reranking choices under a zero-shot constraint. BM25 v2 is the lowest-cost sparse baseline and supplies the shared candidate pool for later systems. MiniLM and BGE-small test whether inexpensive open-weight embedding models improve over that baseline. The prototype MiniLM run tests whether feature engineering at the query level, through ASRS/Likert symptom-language expansion, improves dense reranking. The Gemini run tests a stronger but more expensive LLM reranker over the same first-stage retrieval setting.

\subsection{Sparse Retrieval with Evidence-Aware Rescoring}

Our first system uses BM25 as an unsupervised baseline. Each ASRS question is converted into a bag-of-words query by lowercasing, normalizing punctuation, removing stopwords, and retaining informative terms from the symptom text. Candidate sentences are scored with BM25 over the sentence \texttt{TEXT} field.

To reduce false positives from purely lexical matching, we developed a BM25 v2 pipeline that adds conservative filtering and rescoring heuristics. It rewards lexical cues for first-person and self-report language such as ``I'', ``my'', ``I feel'', ``I struggle'', and ``I can't'', and penalizes sentences dominated by second-person advice, generic ADHD discussion, third-party references, URLs, repeated punctuation, spam, or otherwise low-information text. This produces a sparse retrieval baseline that remains interpretable while prioritizing sentences that are relevant to a user's symptom disclosures.

\subsection{Embedding-Based Reranking}

Our second family of systems follows the first-stage retrieval pipeline described above. We use BM25 v2 as a first-stage retriever and dense semantic similarity as a second-stage reranker. For each ASRS symptom, we retrieve 2000 BM25 candidates and compute cosine similarity between candidate sentence embeddings and symptom query embeddings. We then submit the top 1000 reranked sentences.

We explored two sentence embedding models. The MiniLM run uses \texttt{all-MiniLM-L6-v2} with raw ASRS questions as semantic queries. The BGE-small run uses \texttt{BAAI/bge-small-en-v1.5}, also with raw ASRS questions, over the BM25 v2 candidate pool. These systems are intended to recover paraphrases and semantically related symptom evidence that sparse retrieval may rank too low. Our rationale was to use general-purpose open-weight embedding models as a practical floor for semantic reranking performance rather than relying on costly inference-only encoders.

\subsection{Prototype Query Reranking}

The prototype hybrid run extends dense reranking by changing the query representation. Instead of embedding only the raw ASRS question, it uses a hybrid query prototype that incorporates the question and natural-language Likert-style answer descriptions. For each ASRS item, we embed the answer descriptions corresponding to Likert severity ranks 2--5, compute a severity-weighted centroid over those answer embeddings, average that centroid with the raw ASRS question embedding, and then L2-normalize the resulting vector before cosine reranking. The motivation is that ASRS symptoms are often expressed through severity-graded behaviors rather than a single question wording. For example, a symptom about staying on task may be expanded with prototype statements such as ``I sometimes have trouble staying on task'' or ``I always have trouble staying on task.'' By embedding a richer symptom representation, the reranker can better match sentences describing concrete manifestations of a symptom.

\subsection{LLM Reranking}

The final system uses an LLM reranker after BM25 v2 candidate retrieval. For each symptom, BM25 v2 retrieves 2000 candidates. Because sending all candidates in one request would be inefficient, costly, and difficult to interpret, the candidates are divided into chunks of 10 sentences. Each chunk is sent to \texttt{google/gemini-3-flash-preview} through OpenRouter with the target ASRS symptom, and the model assigns relevance scores and short rationales for the candidate sentences.

This is best described as chunk-level listwise reranking, not a single global listwise reranking over all 2000 candidates. Within each chunk, the model compares multiple sentences jointly. Across chunks, we aggregate the model's per-sentence scores into a global ranking and use the resulting final score to sort candidates. Since chunk scores are not calibrated through a global comparison, we treat the resulting ranking as approximate listwise reranking rather than full-list reranking. This design trades off full-list comparison for tractable API calls, reproducibility, and complete coverage of the 2000-candidate pool.

The reranker prompt instructed the model to act as a careful relevance judge for ADHD symptom evidence retrieval. Given one ASRS symptom and a small batch of candidate sentences, the model scored each sentence from 0 to 100 according to usefulness as direct evidence for the exact symptom. The scoring rubric assigned 90--100 to direct first-person or concrete evidence, 70--89 to strong but slightly indirect evidence, 40--69 to weak, generic, or ambiguous evidence, and 0--39 to unrelated or off-topic sentences. The prompt explicitly rewarded direct first-person or concrete symptom descriptions and penalized generic ADHD discussion, advice, jokes, and weakly related content. It required valid JSON with an \texttt{item\_id}, \texttt{relevance\_score}, and short rationale for each candidate. The completed run judged 36,000 candidate rows across 3600 chunk requests, corresponding to 18 symptoms times 2000 BM25 candidates.

\section{Results}

We first describe our manual top-10 audit, which was conducted before official qrels were released, and then report the official aggregate leaderboard results for our submitted runs.

\subsection{Manual Top-10 Relevance Audit}

To obtain a qualitative signal before official judgments were available, we manually annotated the top 10 sentences for each symptom and each submitted run, for 900 rows total. Labels were binary: relevant or not relevant. The audit used only the target sentence text, ignored \texttt{PRE} and \texttt{POST} context, did not evaluate within-top-10 ordering, and marked unclear or ambiguous cases as not relevant. For duplicate \texttt{(symptom, sentence)} rows selected by multiple runs, we applied a majority-resolution rule: the pair was marked relevant only when its relevant-label share was greater than 0.5; ties were treated as not relevant.

\begin{table}[t]
  \caption{Manual sentence-only top-10 relevance audit used as a qualitative diagnostic before official qrels were released.}
  \label{tab:manual-audit}
  \centering
  \small
  \begin{tabular}{llrr}
    \toprule
    Run & System & Relevant / 180 & Ratio \\
    \midrule
    run1 & BM25 v2 & 27 & 15.0\% \\
    run2 & MiniLM cosine & 53 & 29.4\% \\
    run3 & BGE-small & 40 & 22.2\% \\
    run4 & Prototype hybrid MiniLM & 55 & 30.6\% \\
    run5 & Gemini LLM & 97 & 53.9\% \\
    \bottomrule
  \end{tabular}
\end{table}

The LLM reranker had the highest manually judged top-10 relevance ratio at 53.9\%. The prototype hybrid MiniLM run followed at 30.6\%, MiniLM cosine at 29.4\%, BGE-small at 22.2\%, and BM25 v2 at 15.0\%. These numbers should not be interpreted as official P@10 because they are based on a single conservative manual audit rather than organizer qrels. They nevertheless suggested that the LLM reranker surfaced more sentence-local symptom evidence in the highest-ranked region.

The manual audit also revealed symptom-level differences. Relevance was highest for lexically distinctive or directly expressed symptoms, especially restless/fidgety behavior, difficulty keeping attention during boring or repetitive work, distractibility by noise or activity, careless mistakes, and fidgeting. Relevance was lowest for more context-dependent or polysemous symptoms, including leaving one's seat, waiting one's turn, remembering appointments or obligations, finishing others' sentences, and misplacing things. This supports the hypothesis that sparse retrieval benefits from uncommon terms that are also clinically specific, while broad or ambiguous terms require stronger semantic disambiguation.

\subsection{Official Results}

Table~\ref{tab:official-results} reports the official Task 3 aggregate scores for our runs, with the top leaderboard run in each judgment setting included as a reference point. The official overview originally labeled the final column as nDCG, and the organizers later clarified that this metric is nDCG@1000.

\begin{table}[t]
  \caption{Official Task 3 results for DS-GT runs, with the top leaderboard run in each judgment setting for reference.}
  \label{tab:official-results}
  \centering
  \small
  \begin{tabular}{llrrrr}
    \toprule
    Setting & System & MAP & R-PREC & P@10 & nDCG@1000 \\
    \midrule
    Majority & Top leaderboard & 0.248 & 0.328 & 0.528 & 0.547 \\
    Majority & BM25 v2 & 0.026 & 0.044 & 0.094 & 0.109 \\
    Majority & MiniLM & 0.038 & 0.075 & 0.189 & 0.134 \\
    Majority & BGE-small & 0.023 & 0.058 & 0.122 & 0.112 \\
    Majority & Prototype MiniLM & 0.043 & 0.078 & 0.194 & 0.146 \\
    Majority & Gemini LLM & 0.085 & 0.115 & 0.300 & 0.207 \\
    \midrule
    Unanimity & Top leaderboard & 0.207 & 0.274 & 0.367 & 0.488 \\
    Unanimity & BM25 v2 & 0.040 & 0.057 & 0.083 & 0.116 \\
    Unanimity & MiniLM & 0.048 & 0.065 & 0.144 & 0.135 \\
    Unanimity & BGE-small & 0.026 & 0.053 & 0.083 & 0.100 \\
    Unanimity & Prototype MiniLM & 0.052 & 0.078 & 0.133 & 0.142 \\
    Unanimity & Gemini LLM & 0.080 & 0.121 & 0.217 & 0.188 \\
    \bottomrule
  \end{tabular}
\end{table}

In both judgment settings, the Gemini LLM reranker was our strongest submission. Under majority-vote judgments, it improved over BM25 v2 from MAP 0.026 to 0.085 and from P@10 0.094 to 0.300. Compared with the top majority-vote run, Gemini LLM obtained MAP 0.085 versus 0.248 and P@10 0.300 versus 0.528. Under unanimity judgments, Gemini LLM again had the highest MAP among our runs, with P@10 0.217. Its leaderboard position improved from 14th under majority-vote judgments to 12th under unanimity judgments.

The manual top-10 audit in Table~\ref{tab:manual-audit} is consistent with this official P@10 pattern. Although it was a single-annotator, sentence-only review and should not be treated as an official metric, run5 had the highest manually judged top-10 relevance ratio by a large margin. Run4, the prototype hybrid MiniLM system, was our second-best MAP run in both official judgment settings. Its improvement over the plain semantic rerankers supports the value of expanding ASRS queries with Likert-style examples of symptom expression, especially answer levels 2--5. The BM25 v2 baseline also remained competitive: by MAP it outperformed the BGE-small semantic reranker in both majority and unanimity settings, showing that generic semantic similarity alone did not reliably improve symptom evidence ranking.

These results suggest three patterns. First, LLM reranking produced the strongest early precision among our systems. Second, prototype query expansion was the most effective non-LLM method by MAP. Third, generic dense reranking did not consistently outperform evidence-aware sparse retrieval, indicating that symptom-ranking quality depends on query construction and evidence filtering, not only semantic similarity.

\section{Discussion}

\subsection{Methodological Observations}

During qualitative development, we observed that naive BM25 often retrieved sentences that matched symptom words but did not provide useful evidence about the writer. For example, query terms related to ``trouble'' or ``wrapping up'' could retrieve generic explanations, advice to another user, or metaphorical uses unrelated to ADHD functioning. This motivated the BM25 v2 evidence-aware rescoring, and it also helps explain why the sparse baseline remained competitive despite its simplicity.

Our semantic reranking runs used general-purpose sentence embedding models rather than models explicitly designed or fine-tuned for passage reranking, such as cross-encoders or rankers trained on MS MARCO-style relevance data. This was a development-time choice aimed at establishing a low-compute zero-shot baseline, but it likely limited the effectiveness of the dense rerankers. Similarly, the LLM system should be interpreted as approximate chunk-level listwise reranking rather than a calibrated global reranker. More rigorous listwise methods, including RankGPT-style approaches, may provide stronger comparisons across larger candidate sets.

\subsection{Embedding-Space Diagnostic}

\begin{figure*}[t]
  \centering
  \includegraphics[width=.96\linewidth]{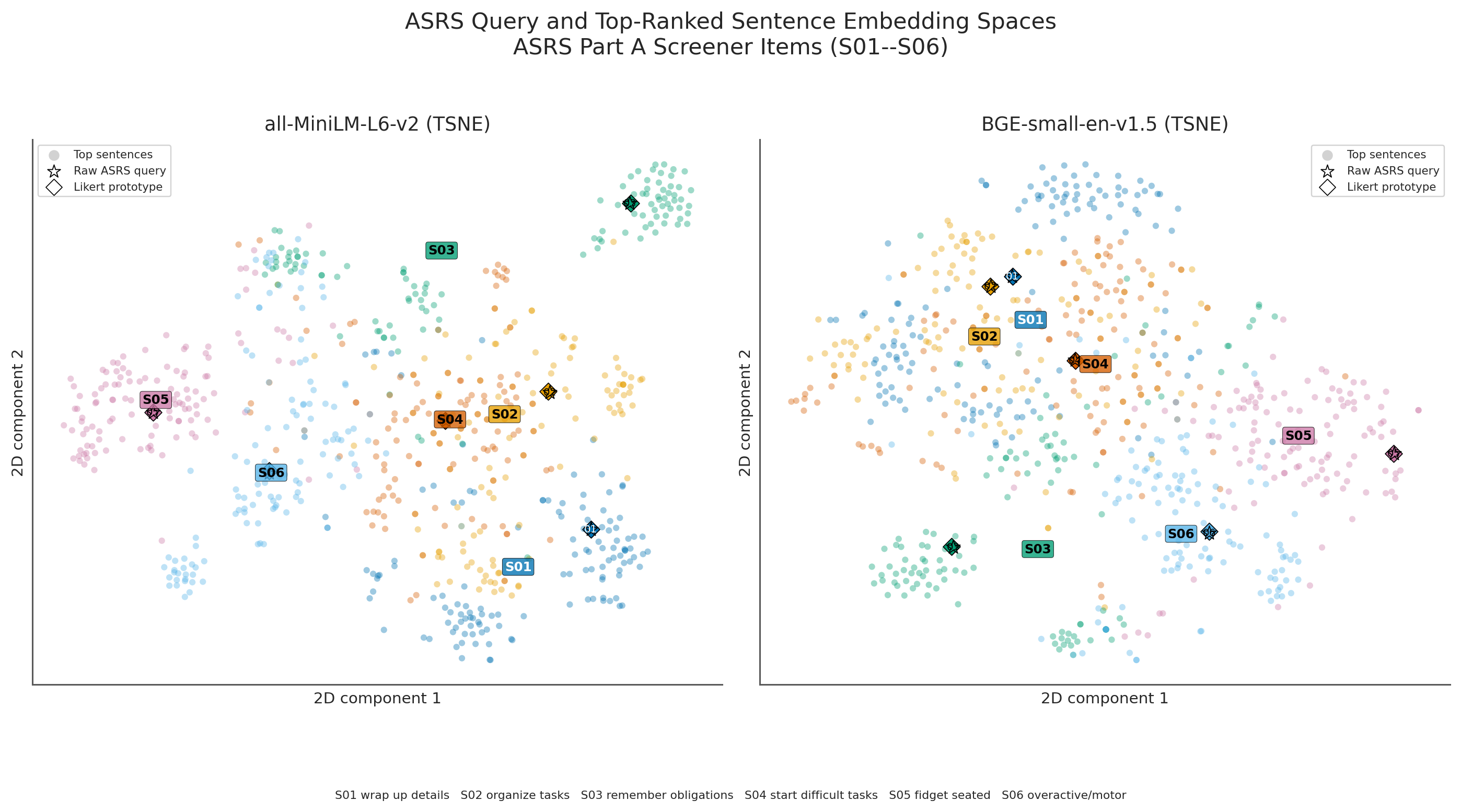}
  \vspace{0.6em}
  \includegraphics[width=.96\linewidth]{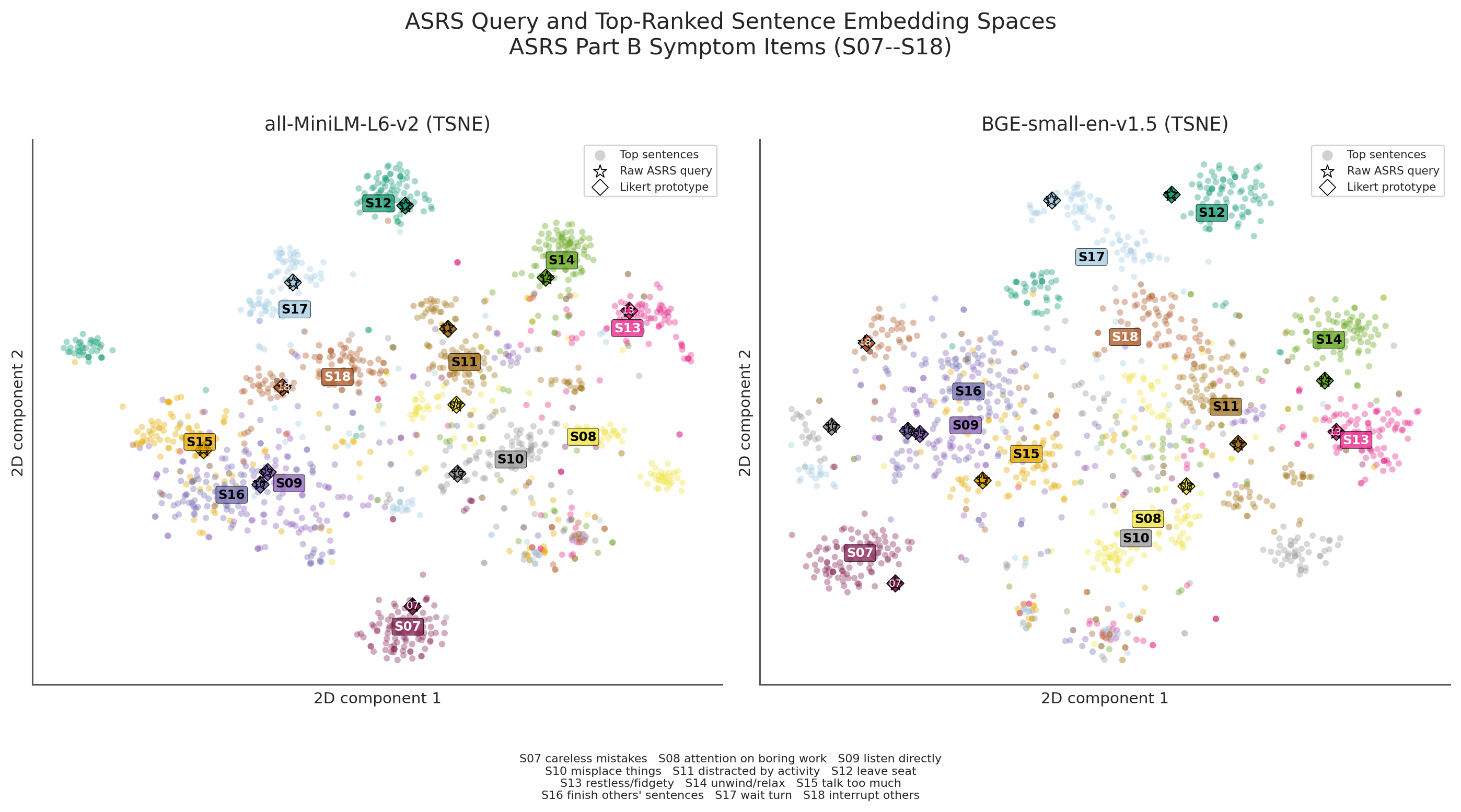}
  \caption{Two-dimensional t-SNE projections of top-ranked sentence embeddings for ASRS Part A and Part B symptoms under MiniLM and BGE-small. Colors denote symptoms, stars mark raw ASRS queries, and diamonds mark Likert-expanded prototype queries. The plot is qualitative and illustrates overlap among symptom regions rather than a formal separability metric.}
  \label{fig:embedding-space}
\end{figure*}

To better understand the behavior of the two dense rerankers, we visualized the top-ranked sentence embeddings for MiniLM and BGE-small using two-dimensional projections, shown in Figure~\ref{fig:embedding-space}. These plots should be interpreted qualitatively, since t-SNE preserves only part of the original embedding geometry. Nevertheless, they help explain why generic dense similarity did not uniformly improve ranking quality.

For the ASRS Part A screener items, MiniLM places S02, S04, and nearby S01 evidence in overlapping regions, reflecting the conceptual similarity between organizing tasks, initiating difficult tasks, and finishing project details. BGE-small shows an even tighter grouping of the corresponding query and prototype markers for S01, S02, and S04, suggesting that these executive-function items are difficult to distinguish with raw ASRS question embeddings alone. In contrast, S03, which concerns remembering appointments or obligations, is more clearly separated. S05, related to fidgeting while seated, also forms a more distinct cluster. S06, which reflects feeling driven or compelled to be active, is separated from the task-organization items but appears closer to the fidgeting-related region, consistent with its motor-restlessness content.

For the Part B symptom items, overlap is more pronounced, especially in the BGE-small projection. MiniLM shows somewhat clearer symptom-level structure, while BGE-small compresses several symptoms into nearby regions. This pattern is consistent with the official results, where BGE-small underperformed the BM25 v2 baseline by MAP in both judgment settings. Conversational-context symptoms such as S09, S15, and S16 appear close together, which is expected because listening, talking too much, and finishing others' sentences can share similar social-interaction language. S08 and S10 also appear near one another, suggesting that attention lapses and misplacing objects may be expressed through related everyday-functioning descriptions. By contrast, S12, S13, and S14 form more distinct regions, and S07 appears relatively separable. Overall, the visualization suggests that semantic reranking is most useful when symptom language occupies distinguishable regions of the embedding space, but less reliable when closely related ASRS items share broad functional or conversational language.

\subsection{Limitations and Future Work}

This analysis has several limitations. The manual audit was single-annotator and sentence-only: it ignored \texttt{PRE} and \texttt{POST} context, collapsed unclear cases into not relevant, and did not evaluate ordering within the top 10. The LLM reranker used chunk-level scoring rather than global listwise comparison, so candidates from different chunks were not directly compared in the same model call. In addition, the first-person heuristics may underweight relevant evidence that is implicit, contextual, or not expressed through explicit self-report lexical cues. Finally, the official results reported here are aggregate scores, so per-symptom behavior remains unresolved.

In future work, we would like to compute per-symptom effectiveness to identify which ASRS items are most difficult for our systems and inspect representative failures. We would also like to run an automated LLM-assisted relevance review over ranked sentences for each symptom. Such labels would be treated as exploratory rather than authoritative since they would be produced by an AI judge. Their purpose would be to summarize likely error patterns, such as lexical ambiguity, insufficient self-reference, third-party discussion, or symptom descriptions that require broader context. These analyzes could support symptom-specific query expansions, improved filtering for ambiguous symptoms, and selective LLM reranking over smaller candidate pools. We would also like to evaluate reranking models designed specifically for retrieval, including cross-encoder rerankers and stronger listwise approaches such as RankGPT-style methods, to compare them against the general-purpose embedding and chunk-level LLM reranking strategies used in this work.

\section{Conclusion}

We submitted five unsupervised systems for eRisk 2026 Task 3, spanning sparse retrieval, dense reranking, query-prototype reranking, and LLM reranking. These systems explore a practical range of retrieval strategies for a first-edition task with no annotated ADHD training data. Official results confirm that the Gemini LLM reranker was our strongest submission, while prototype-based MiniLM reranking was the strongest non-LLM alternative by MAP.

\begin{acknowledgments}
We thank the Data Science at Georgia Tech (DS@GT) ARC group for their support and the eRisk 2026 organizing committee for designing and running this task. Computing resources were provided by the Georgia Institute of Technology in Atlanta, Georgia, USA, through access to the Partnership for an Advanced Computing Environment (PACE)~\citep{pace}.
\end{acknowledgments}

\section*{Declaration on Generative AI}
During the preparation of this work, we used Claude (Anthropic) and ChatGPT (OpenAI) to check grammar and spelling, improve flow, and format LaTeX. After using these tools, we reviewed and edited the writing as needed and take full responsibility for the publication's content.

\bibliography{task3_erisk2026}

\end{document}